\pdfoutput=1

\documentclass[11pt]{article}

\usepackage{acl}

\usepackage{times}
\usepackage{latexsym}
\usepackage{pifont}

\usepackage[T1]{fontenc}

\usepackage[utf8]{inputenc}

\usepackage{microtype}

\usepackage{inconsolata}

\usepackage{inconsolata}
\usepackage{multirow}
\usepackage{graphicx}
\usepackage{longtable}

\newcommand\ie{\textit{i.e.}\ }
\newcommand\eg{\textit{e.g.}\ }

\title{Principles from Clinical Research for NLP Model Generalization}

\author{\bf Aparna Elangovan$^1$, 
  \bf Jiayuan He$^{2,1}$,
  \bf Yuan Li$^{1,2}$,
  \bf Karin Verspoor$^{2,1}$\\
  $^1$The University of Melbourne, Australia\\
  $^2$RMIT University, Australia \\
  {\tt aparnae@student.unimelb.edu.au} \\
  {\tt \{jiayuan.he, yuan.li, karin.verspoor\}@rmit.edu.au}}

\begin{document}
\maketitle

\begin{abstract}

The NLP community typically relies on performance of a model on a held-out test set to assess  generalization. Performance drops observed in datasets outside of official test sets are generally  attributed to ``out-of-distribution'' effects.  Here, we explore the foundations of generalizability and study the  factors that affect  it,  articulating  lessons from clinical studies. In clinical research,  generalizability is an act of reasoning that depends on  (a) \textit{internal validity} of experiments to ensure controlled measurement of cause and effect, and (b) \textit{external validity} or transportability of the results to the wider population. We demonstrate how learning spurious correlations, such as the distance between entities in  relation extraction tasks, can affect a model's internal validity and in turn adversely impact  generalization.  We, therefore, present the need to ensure internal validity when building machine learning models in  NLP. Our recommendations also apply to generative large language models, as they are  known to be sensitive   to even minor semantic preserving alterations. We also propose adapting the idea of  \textit{matching} in randomized controlled trials and observational studies to  NLP evaluation to measure causation.
\end{abstract}

\section{Introduction}
What factors lead to poor generalizability of models? To understand causes of   non-generalization, the notion of generalizability needs to be first clearly defined. One definition of generalizability, in the context of general machine learning, is the ability of a model to perform well on data \textit{ unseen during training}, but \textit{drawn from the same distribution or population} \cite{google_2022}. Data \textit{unseen during training}  clearly refers to data that is not part of training data, and is arguably uncontroversial to support sound evaluation.  
However, the requirement of data \textit{drawn from the same distribution or population}  warrants more scrutiny.

It is important to consider the impact of data distribution on generalization, yet
this is very challenging for natural language data. 
In statistics, the concept of distribution indicates ``the pattern of variation in a variable or set of variables in the multivariate case'', and thus describes the frequency of values of an observed variable \cite{WildTHECO}. 
Distribution or frequency of observed variable values can be challenging notions to  meaningfully adapt to 
high dimensional, multivariate, and high variability text data. 
Currently, no formal definition of in-distribution or out-of-distribution (OOD) texts, or how to detect them, exists for NLP \cite{arora-etal-2021-types}. Where distributions are considered, simple surface linguistic characteristics or lexical-level distributions are emphasized~\cite{verspoor2009textual}.

Despite the lack of  a comprehensive formal definition of OOD in existing NLP literature, OOD has become the most commonly cited reason for generalization failure when a model performs poorly outside an official test set. Due to the black-box nature of deep learning models, it is increasingly difficult to demonstrate if a model has established a robust decision-making process that is generalizable to unseen data. As a consequence, 
generalization failures are typically ascribed 
 to external factors, i.e.\ those extraneous to model development practices, and primarily to shifts in data distribution, or OOD. 

Recent studies have emerged  showing that the robustness of a model can be undermined by inadvertent errors made during development of a model. For example, it has been shown that  data leakage from training into test splits can lead to inflated test results~\cite{elangovan2021memorization}. Other works have pointed out spurious correlations  in various benchmark datasets that may 
be leveraged by deep learning models to achieve inflated performances on test sets, while having poor generalization capability to real-world settings~\cite{gururangan-etal-2018-annotation, mccoy-etal-2020-berts, shinoda-etal-2022-look}. 

In this paper, we  hope to draw  attention to the causes of NLP model  generalization failures, especially those internal factors that are part of the model development process (\eg   preparation of training data). We argue that the external validity of a model 
should only be examined if the internal validity of the model 
is established. 
We start with a case study, showing how generalization failures can be caused by internal factors -- the model has learned surface patterns in training data. We then propose that a pragmatic notion of model generalizability in the NLP domain can be established through borrowing and adapting practices from a domain seemingly far afield -- clinical research.

The contribution of this paper is two-fold. First, we show how OOD may not be the sole cause of generalization failures, via a relation extraction task, highlighting the need for intrinsic investigation of why models fail. Second, we propose to categorize the  causes of generalization failures in  NLP models, drawing inspiration from clinical studies. Our work provides guidance on how to more systematically analyze generalization failures and adapt the principles behind randomized controlled trials for NLP model evaluation.

\section{Relation extraction case study}
Through a relation extraction case study over two data sets, we demonstrate that poor generalization in real-world application can result from ineffective modelling, \ie learning of superficial surface patterns, rather than data distribution shift. For completeness, we also study a  popular benchmark dataset on natural language inference (NLI) task.

\subsection{Approach}
Due to the black-box nature of deep learning models, it is difficult to interpret the  underlying basis for model predictions. 
Inspired by recent works in explainable NLP models, such as LIME~\cite{ribeiro-etal-2016-trust}, we employ interpretable surrogate models to examine the behavior of deep learning models. Specifically, assume a dataset $\mathcal{D}=\{\langle x, y\rangle\}$, where $x$ represents the input sequence of words and $y$ represents the ground-truth label for $x$. We train two surrogate models: a model $S_{g}$ that is fit on the dataset $\langle x_u, y\rangle$ and another model $S_{\hat{m}}$ that is fit on the dataset $\langle x_u, \hat{y}_m\rangle$, where $x_u$ represents $x$ via a representation technique $u$ and $\hat{y}_m$ is the prediction of the deep learning model $B_m$ to be examined, e.g.\ a BERT-based model~\cite{devlin-etal-2019-bert}. For $u$, we adopt a vector of surface patterns.

We hypothesize that a strong correlation between the predictions of the surrogate $S_{\hat{m}}$ and the corresponding main model predictions indicates that the underlying model $B_m$ has relied on the surface patterns in $x_u$, and that the model's predictions may not be reliable. The idea here is that if a surrogate model can reproduce the behavior of a comparator model with high fidelity, then that surrogate model is a good approximation of that comparator model and hence there is no evidence that the comparator model has learned anything more than the patterns captured in the surrogate model. This also follows from the principle of Occam’s razor related to the law of parsimony \cite{epstein1984principle, felsenstein1983parsimony}. We interpret the correlation between the predictions of $S_{\hat{g}}$ and the ground truth labels as the indicator of the extent to which the surface patterns $x_u$ are present in the underlying data. A strong correlation indicates the weakness in the dataset itself and how these patterns can be exploited to achieve highly accurate predictions without deeper linguistic comprehension. We use  Cohen's Kappa $\kappa$ to measure  correlations.

\subsection{Datasets}
We use the following datasets:
\begin{itemize}
\setlength\itemsep{.05 mm}
    \item \textbf{PTM-PPI (PTM)} is sampled from PubMed abstracts \cite{Elangovan2022-vl} for relation extraction (REL) task, annotated with 6 types of post-translational modification relationship between two proteins. Out of the 6 positive classes, we only consider the class ``phosphorylation'' since only this class has a sufficient number (> 100)  of training samples. Consequently, the dataset used has 2 classes: ``phosphorylation'' class and the negative class. 

    \item \textbf{ChemProt (CHM)} is sampled from PubMed abstracts, annotated with 5 types of protein–chemical relationships \cite{Krallinger2017OverviewOT} for REL task. The dataset contains 6 classes in total: 5 positive classes and the negative class.

    \item \textbf{SNLI (SNL)} is a NLI task dataset with 3 classes \cite{bowman-etal-2015-large}. 
\end{itemize}
 For CHM and PTM, we fine-tune a BioBERT model~\cite{BioBERT}. For SNLI dataset, we fine-tune a BERT \cite{devlin-etal-2019-bert} model.

\subsubsection{Generalization  data sets}
 
 To  understand the impact of generalization behavior  of a model that has relied on spurious factors, beyond the test set, we selected the REL tasks  due to the availability of data on PubMed. We select a random subset of PubMed abstracts to create a ``generalization'' set.  We apply the fine-tuned BioBERT model to generate the predictions for the ``generalization'' set. From this generalization set, we randomly sample from  the top 25 percentile   high confidence predictions to form set GH for each class \cite{Elangovan2022-vl} and report results across 10 runs. Selecting only the high confidence predictions follows \citet{hendrycks2017a}, which demonstrated that the prediction probabilities of OOD samples tend to be lower than those of correct samples. Table~\ref{tab:datasetatrib} summarizes the datasets.

\begin{table}[]
    \centering
    \begin{tabular}{llr}
    \hline
         \textbf{Dataset} & \textbf{Split (Label)} & \textbf{\# Pos / \# Neg} \\
     \hline
          \multirow{4}{*}{PTM}& TR (GT) & 139 / 1116 \\         
          & TS (GT)  & 44 / 308 \\
          & TS (MP) & 24 / 328\\
          & GH (MP)  & 250 / 5000  \\

        \hline
          \multirow{4}{*}{CHM} & TR (GT) & 4172 / 2265 \\
          & TS (GT) & 3469 / 2275 \\
          & TS (MP) & 3726 / 2018\\
          & GH (MP)  & 7500 / 2500 \\
       
       \hline
        \multirow{3}{*}{SNL} & TR (GT) & 366374 / 182764 \\
          & TS (GT) & 6605 / 3219 \\
          & TS (MP) & 6462 / 3362\\
         \hline
    \end{tabular}
    \caption{Summary of data sets. TR: official training set; GT: ground-truth labels are used; TS: official test set; MP: model  prediction labels, GH: generalization set.}
    \label{tab:datasetatrib}
\vspace{-5mm}
\end{table}

\subsection{Surrogate models and surface patterns}
We employ two explainable surrogate models: 
\begin{itemize}
\setlength\itemsep{.05 mm}
    \item \textbf{Multinomial Naive Bayes (NB)}: The Multinomial Naive Bayes approach represents input samples using simple surface patterns (n-grams, $n=1$). 
    To avoid over-fitting, we select the top $k$ most commonly n-grams per class. Hence, the maximum number of n-grams that the NB model uses is $k\times$ number of classes. We set $k$ as 100 in the experiment.
    \item \textbf{Naive Bayes + Decision Tree (NB-T)}: Here we use model stacking, where a Decision Tree is stacked on top of a NB model. In NB-T, the prediction of NB is used as one feature input in the subsequent decision tree. Additionally, handcrafted rules, detailed in section~\ref{sec:handcraftedrules}, are used as surface pattern features in the decision tree. To void over-fitting and to allow the decision process to be explainable, we restrict the tree depth to $<=4$. 
\end{itemize}
\subsubsection{Crafted surface pattern features}\label{sec:handcraftedrules}
To study whether the BioBERT-based model has relied on distance-based surface patterns for relation extraction, we  manually analyze BioBERT's predictions on the PTM task and  identified  4  surface patterns  that potentially explain the model predictions. We then use these  hand-crafted surface patterns to represent the inputs to a surrogate  model and verify how well the predictions of the surrogate  model 
correlates with BioBERT's predictions (which is given the full-text input).  The 4 surface patterns are:
\begin{itemize}
 \setlength\itemsep{.05 mm}
    \item \textbf{Percentage count of participating entities (E1C and E2C) :} Given an input sentence $s$ and a relation $\langle E1, Rel, E2\rangle$ in $s$, this feature captures the percentage of the total tokens in $s$ corresponding to each entity. For instance, for the input \textit{``GENE\_A interacts with CHEMICAL\_C and binds to CHEMICAL\_C and CHEMICAL\_D"}  and  the relation $\langle$CHEMICAL\_C, Rel, GENE\_A$\rangle$, the features E1C and E2C are $\frac{2}{10} * 100 = 20.0$ and  $\frac{1}{10} * 100 =10.0$, respectively (input contains 10 words, E1 occurs twice and E2 once).
    \item \textbf{Length of shortest span containing the participating entities and a given trigger word T (LSS\_$\langle$T$\rangle$):} This feature represents the length of the shortest span containing the two entities and a specified trigger word. For instance, an input \textit{``\underline{GENE\_A interacts with CHEMICAL\_C} and binds to CHEMICAL\_C and CHEMICAL\_D"} and  $\langle$CHEMICAL\_C, Rel, GENE\_A$\rangle$ the length of the shortest span containing the trigger word ``interacts", LSS\_interacts, is 4, whereas LSS\_binds is 6.
    \item \textbf{Length of shortest span that contains the entities and any trigger word (LSS):} For instance, given input \textit{``\underline{GENE\_A interacts with CHEMICAL\_C} and binds to CHEMICAL\_C and CHEMICAL\_D"} and relation $\langle$CHEMICAL\_C, Rel, GENE\_A$\rangle$, the shortest span has length 4.    
    \item \textbf{Fraction of sentences containing participating entity pair (SPC):} This feature represents  the normalized count of sentences containing the entity pair. For instance, if the input text contains $s_n$ sentences and only $k$ sentences contain both entities E1 and E2 in $\langle E1, Rel, E2\rangle$, then this feature would be $\frac{k}{s_n}$.
\end{itemize} 
For SNLI, we use the spurious factors reported by \citet{gururangan-etal-2018-annotation}, such as the length of the hypothesis and presence of negation.

\subsection{Results}\label{sec:results}
Table~\ref{tab:ptmsurfacepattern} reports the 
Kappa 
correlation between \textbf{(a)} the  surrogate models and ground truth; and  \textbf{(b)} the surrogate models and  fine-tuned model's prediction. 
For the PTM corpus, NB-T achieves better correlation with ground-truth, compared to NB, in all settings, demonstrating that the hand-crafted surface patterns are more likely than n-grams along to be influencing model predictions. In addition, when ground-truth labels are used as targets, i.e.\ TS (GT) \textit{vs} $S^{test}_g$ predictions and TR (GT) \textit{vs} $S^{Train}_g$ predictions, NB-T correlation   $\kappa$ is 0.55 and 0.54 respectively, indicating that similar surface patterns exist in both the training and test sets of ground-truth labels. This is not surprising given that the test and train sets were obtained using a random split from a single dataset. 

To examine if the fine-tuned BioBERT model has relied on those hand-crafted surface patterns, we compare the correlation between NB-T on the test set when fitting to the ground-truth labels, TS (GT), versus the BioBERT model's predictions, TS (MP). We see that the $\kappa$~correlation increases drastically from 0.55 -- weak correlation, to 0.73 -- moderate correlation (based on \citet{McHugh2012-pc} ranges), when the target labels are replaced with BioBERT predictions (cf.\ trees in Appendix~\ref{App:PTM-PPI dataset Decision-tree}). This demonstrates that NB-T  using handcrafted features is more correlated with BioBERT's predictions compared to ground-truth labels themselves, increasing the 
evidence that the BioBERT model may rely 
 on these features. This phenomenon is further exacerbated on the GH set, where NB-T achieves  $\kappa$~correlation of 0.85 - strong correlation when fitting to BioBERT's predictions. In fact, \citet{Elangovan2022-vl} report that only 6  out of 30 (20\%)  of  the phosphorylation predictions turned out to be accurate when the high confidence predictions were randomly sampled and verified by experts, compared to test set precision of 62.5\%. The drop in precision, compared to test set performance, \textit{may be} 
explained by the model relying on these surface patterns rather than broadly generalizable features.%
\begin{table}[]
\setlength\tabcolsep{1.2pt}
    \centering

 \begin{tabular}{llrrr}
 \hline
\textbf{DS (L)} & \textbf{SM} & \textbf{PTM $\kappa$ } \hspace{1em}   & \textbf{CHM $\kappa$} \hspace{2em} & \textbf{SNL $\kappa$} \\
\hline 

\multirow{2}{*}{TR (GT)} &  NB    &   0.33 \hspace{2em}  & 0.45 \hspace{3em} & 0.25  \\

&  NB-T  &   0.54 \hspace{2em}  & 0.46 \hspace{3em} & 0.27 \\

\hline
\multirow{2}{*}{TS (GT)} &  NB    &   0.26 \hspace{2em} & 0.48 \hspace{3em} & 0.29   \\

&  NB-T  &   0.55 \hspace{2em}  & 0.48 \hspace{3em} & 0.33  \\

\hline
\multirow{2}{*}{TS (MP)} &  NB    &   0.25 \hspace{2em} & 0.50 \hspace{3em} & 0.32 \\

 &  NB-T  &  \textbf{0.73} \hspace{2em}  & 0.51 \hspace{3em} & 0.34  \\
 
 \hline
                         
\multirow{2}{*}{GH (MP)} &  NB     &  0.77 (0.3) & 0.44 (0.002)&  -  \\
                         &  NB-T  &  \textbf{0.85} (0.3)& 0.44 (0.002) & - \\

\hline
\end{tabular}
    \caption[Correlation using surface patterns.]{Surrogate model correlations on dataset (DS) and the target label (L): The surrogate model (SM) NB-T correlates better with BioBERT's prediction than the ground truth labels for the PTM dataset. For GH set,  we report standard error for 10 runs ($\frac{\sigma}{\sqrt{n}}$, where $n=10$).  The p-value for Cohen's-$\kappa$ is less than 0.05.}
    \label{tab:ptmsurfacepattern}
\vspace{-6mm}
\end{table}

In the CHM dataset, distance based
surface patterns do not  seem to be a stronger predictor than n-grams, as shown in Table~\ref{tab:ptmsurfacepattern}. All the surrogate models have a correlation between 0.4 and 0.5 indicating weak correlation \cite{McHugh2012-pc}. 

For SNLI,  the surrogate models achieve minimal correlation between 0.21 and 0.39 \cite{McHugh2012-pc} indicating there are potentially other features required to improve the surrogate model.  The hypothesis length, as reported by \citet{gururangan-etal-2018-annotation}, is  one of the key features that NB-T also identifies, see details in Appendix~\ref{app:sec:SNLI}. 

\textbf{OOD is NOT always a sufficient explanation for generalization failures.} Given the broad and generic definition of OOD,  almost any sample can be categorized as OOD. It  is difficult to counter the OOD argument,  as there is  no comprehensive approach to establishing that a given instance is in-distribution \cite{arora-etal-2021-types}.  While in the case of CHM, we were unable to detect clear surface patterns, in the case of the PTM dataset,  BioBERT \textit{appears to heavily rely on  surface patterns} reflected in our handcrafted distance-based patterns. The strong correlation between the surrogate model and the BioBERT-based model prediction points to the model potentially 
relying on such surface patterns, which will undoubtedly hinder the model's generalizability to external datasets.
 Therefore, before concluding OOD as a potential cause of generalization failure, we need to ensure that spurious correlations are NOT the source of a model's high performance. There are cases where some surface patterns might be reasonable for some tasks, we discuss such cases  in detail in Section~\ref{sec:unreliablesuracepattern}. While we acknowledge that correlation does not necessarily imply causation, under certain conditions it may indeed  \cite{gardner2000correlation}. 
 Our results above and several other works suggest that spurious factors might be enabling models to achieve high performance \cite{gururangan-etal-2018-annotation, mccoy-etal-2019-right}, but further (difficult to design) tests would be required to unambiguously establish that \textit{the cause} of high performance is indeed spurious correlations.

Detecting spurious correlations is non-trivial, while  simple surrogate models can detect dominant surface patterns provided we know what to look for apriori. Hence, our  approach of using surrogate models has two main challenges: \textbf{a)} it requires good handcrafted features, and \textbf{b)} it assumes only a few  dominant patterns exist.   Deep learning models, such as transformers, that can potentially learn thousands of low frequency surface patterns (that may not be detectable by simple models), while these patterns may also be present in the test set, leading to inflated test performance.

\section{Foundations of generalizability}
The core idea behind generalizability is that the conclusions drawn, or a model inferred,  from a sample can be applied to a wider population. In this section, we discuss some of the foundations of generalizability from clinical studies and why  generalization failures cannot be solely attributed  to out-of-distribution factors. Identifying the cause of generalization  failures requires several components to be well-defined ahead of the experiments, which are discussed in detail in this section.

\begin{figure}[h!]
    \centering
    \includegraphics[width=0.9\linewidth]{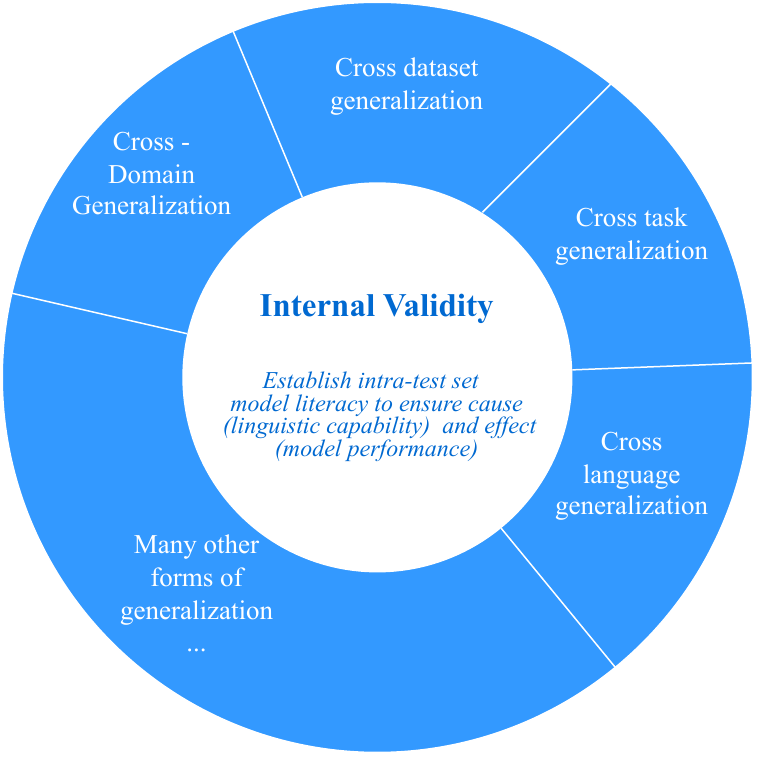}
    \caption[Internal validity]{Internal validity is a mandatory precursor for any form of external generalization, including cross dataset generalization. Internal validity is required to ensure that the model has learned 
    core linguistic strategies to solve the task within the context of the test set.}
    \label{fig:internalvalidity}
\vspace{-5mm}
\end{figure}

\subsection{The notion of generalizability}
Clinical studies aim to answer questions such as \textit{``is this drug treatment effective?"}, and critically rely on generalizability to establish meaningful evidence~\cite{rothwell2005external,guyatt2011grade8,schunemann2013non}. 
A core element of a clinical study is the specification of  
the study population, referring to a subset of the population selected for research;  it is impossible to study the entire population \cite{Kukull2012-dr}. The implicit assumption is that  conclusions  drawn from the sample are applicable to the population, requiring the population boundary to be defined.  This boundary 
depends on the aim of the study, can include various factors, including country, insurance memberships or disease status \cite{Kukull2012-dr}. For instance, for a study that investigates the effects of a drug on a disease, the relevant population would typically be all the people with that disease. To ensure that any conclusions from the study using the samples drawn from the population are confidently generalizable  to the entire relevant population,  \textit{intrinsic} and \textit{extrinsic validity} of experiments \cite{guyatt2011grade2} must be established.   We detail these concepts below.

\subsection{Study Aim}
In machine learning or more specifically NLP, the aim of a study might seem obvious, for example, to establish whether one model is better than the other for a given task  according to a chosen metric. However, even 
such a straightforward research goal 
can be ambiguous, as the conclusion drawn depends on the dataset used to evaluate the models. Hence, if the study aim is well-defined or constrained, e.g.\ referring to performance on a benchmark such as GLUE  \cite{wang-etal-2018-glue}, then the objective is clear, allowing the conclusions to be contextualized. In real-world settings, the objective is usually to know if the model can meet certain performance objectives, e.g.\ will the model's predictions be at least 70\% accurate when deployed.  As a consequence, it becomes pertinent to define the population boundary or the context to ensure that the model performance is optimal for the task it is designed for in that broader context. Hence, defining the aim of the study requires careful definition of the target  population.

\subsection{Defining populations}
In the context of machine learning, if a test set is a sample that effectively represents a population, then the conclusions based on the test set should apply to the entire population. For instance, if the conclusion is that the model has an accuracy of 90\% based on the test set, it should mean that when the model is applied to the entire population,  the prediction accuracy ideally should also be ${\sim}90\%$.  

In NLP, for the datasets used to benchmark model performance, such as GLUE \cite{wang-etal-2018-glue}, the broader ``population'' corresponding to 
these datasets is  not clearly defined, and hence it is difficult to define the boundary or context of generalizability, \ie the population that these results are meant to apply to or when data is  out-of-distribution (OOD). As a result, when a model performs poorly on a different test set, 
an explanation that the data is OOD is 
insufficient unless the notion of distribution is clearly defined. We cannot  simply attribute poor performance to OOD data.  Model performance actually depends on \textbf{(a)} what the model has learned, and \textbf{(b)} how effective a test set is in measuring its performance.   Moreover, OOD data should ideally have lower model confidence scores \cite{hendrycks2017a}. Thus, claims of OOD as an explanation for poor performance at the very least require the context of population or distribution to be clearly defined. 

When we define a new task, 
we implicitly define the population boundary for the task, such as through the use of data cards \cite{datacards}.
For instance, if our task is to analyze the sentiment of IMDB movie reviews,
IMDB movie reviews are the population of texts that this model applies to. However, defining the boundary of this population is not trivial. It may require additional constraints around language, written vs.\ spoken style, 
and more precise specification of the 
domain, such as -restaurant vs.\ movie reviews. 

An effective, well-informed boundary should consider key factors that can impact the performance of a model in a real-world scenario and constrains the problem space so that the samples can be drawn from the population that the model is meant to serve. Hence, defining the population boundary is a mandatory precursor to collecting  training and test data to ensure that the collected samples are   representative of the population.

\subsection{Internal validity}\label{sec:internalvalidity}

Internal validity is crucial to ensure that the measurement of the relationship between \textit{cause and effect} is not affected by 
spurious correlations or bias in the data \cite{Delgado-Rodrguez635}. This particularly affects what we can infer from a gain in model performance. 

To understand  \textit{internal validity}  in the NLP context, 
consider a hypothetical example of  a customer sentiment analysis text classification task. To collect data, we may randomly select 500 customer emails  from organization A (org-A), and another 500   from organization B (org-B). Let's assume that org-A generally provides better customer service than org-B, and that the samples from org-A contain a signature marker, ``FROM-ORG-A''. Say that a deep learning model that requires no feature engineering has over 90\% accuracy, while another simpler model based on carefully curated  semantic features achieves a performance of 75\%. The software code is well tested, and the researchers also perform statistical significance tests and conclude that the deep learning model is better than the simpler model. However, the deep learning model has, in fact, relied on the signature ``FROM-ORG-A'' as a key indicator for positive labels, while the simpler model relies on the presence of words such as ``great'', ``mediocre'', etc.\ to differentiate between classes. 
Is the conclusion that the deep learning model is better than the simple model at customer sentiment analysis internally valid?
The same parallels can be drawn from the case study of the PTM dataset discussed above, where the model seems to have relied on spurious correlations. Hence, the performance on a test set need not indicate that the model has the \textbf{basic   linguistic task level literacy}  even within the limited scope of the test set, as depicted in Figure~\ref{fig:internalvalidity}.

Internal validity of experiments can also be affected by factors such as sample selection and instrumentation \cite{Wortman1983}.
Internal validity is to ensure that the study and the conclusions are valid within the context of the experiment, where the cause (model has learned the right aspects of the language) and effect (model's performance) is fairly evident. Experimental errors  such as bugs in the code, issues with test/training split resulting in data leakage \cite{elangovan2021memorization}  are obvious examples of errors that invalidate results.  Factors such as dataset bias, data splits and test data issues affect reproducibility of experiments \cite{gundersen2023sources} also affect internal validity.
The internal validity can also be affected by the selection of participants in the study \cite{Patino2018-pm}. The participants of a study in NLP 
can be construed as  the data and any human annotators. Lack of careful consideration of details such as training or test sample size, sample selection criteria etc.\ can make the study lack internal validity. 

Performance gains made by large language models can be misunderstood as natural language understanding \cite{stochasticpaper}. 
\citet{shortcut2020}  also emphasize the need to differentiate the capability required to perform on a dataset cf.\  the underlying capabilities of a model. Robustness of experimental design, data selection criteria,  well-tested code, careful train-test split, effective test sets, statistical analysis, etc.\  are core aspects of  internal validity in NLP.

\subsection{External validity}
External validity, associated with \textit{transportability} of results from  samples to the wider population, is a heavily debated topic even in clinical research, whereas internal validity is a more established concept in clinical studies~\cite{Tipton2014HwWGeneralizable, yarkoni_2022, Degtiar2023}. If a study is not internally valid, then external validity is irrelevant \cite{Patino2018-pm}. Assuming study results are internally valid,  whether the  conclusions of a study are generalizable or transportable to the  wider population 
depends on the ability to separate  ``relevant" from ``irrelevant" facts for the study. 
Importantly,  well-designed population-based studies can minimize the risk of \textit{selection factors with unintended  consequences} on study results \cite{Kukull2012-dr}.  

Spurious correlations in training data affect internal validly as the model is set up to learn irrelevant facts, while training samples that do not sufficiently represent the underlying population affect external validity~\cite{Delgado-Rodrguez635}.  In clinical studies, conclusions drawn from the sample may lack external validity when there are differences between study samples and the target populations, such as subject characteristics or hospital procedures \cite{Degtiar2023}. For an equivalent example  in NLP, consider sentiment analysis. A trained model with a set of words such as ``good'' associated with positive sentiment may be internally valid, but may fail to perform well on the wider population when it encounters newer terms such as ``heartwarming''. 

In NLP, even though there is no single comprehensive formal definition of OOD   \cite{arora-etal-2021-types}, conceptually generalization challenges stemming from OOD  can generally be considered external validity challenges. OOD can be a result of  domain or distribution shift, due to languages or tasks differing from training data \cite{hupkes2023stateoftheart}, or even samples from adversarial attacks \cite{omar2022robust}.

\section{Discussion}

\subsection{Is the generalization failure due to internal or external validity?}
Attributing the right cause of failures enables us to take the most effective corrective action, hence separating internal vs.\ external factors is important, given internal factors are far more controllable than external ones. Ensuring internal validity requires that we understand cause and effect of a model's performance. This in turn forces researchers to analyze the data,  investigate training methods that are robust against issues in the training data such as noisy labels and spurious correlations. High performance on the test set is clearly not sufficient to ensure that the  model  is capable of solving the task it is trained for, as similar spurious correlations can exist in both training and test sets. Training data is rarely perfect, as it can contain many problems reflecting annotator bias, incorrect or noisy labels~
\cite{mccoy-etal-2019-right, gururangan-etal-2018-annotation}. 

Inspired by prior works  that point to the contributors of poor internal validity, as described in Section~\ref{sec:internalvalidity}, we propose  the following checks to ensure internal validity:
\begin{enumerate}
 \setlength\itemsep{.03 mm}
    \item How many spurious correlations or noisy labels are present in training and/or test data?
    
    \item How diverse is the training data, and is it sufficient in volume to learn the right features for a given task?
    
    \item How robust is the training procedure against  spurious correlations or noisy labels?

    \item Were  model explainability analyses able to identify the model's reliance on spurious correlations?
    
    \item  Are experiments well-designed and reproducible?

    \item How  effective is the test set  in verifying what the model has learned and/or weaknesses in the model?
\end{enumerate}

Questions 1-4 can be difficult to answer accurately in practice due to deficiencies in the current set of tools and technologies available to analyze the large volumes of   data for surface patterns. They may rely on domain knowledge. It may simply be expensive to collect more training data. These challenges are compounded by the fact that neural networks are difficult to understand. 

Good test sets, on the other hand,  provide a pragmatic way to understand the capabilities of a model, even without access to the underlying model architecture.  Ideally, the size of the randomly sampled test set should be sufficiently large, as a large  sample size is much more likely to representative of true performance than a smaller one \cite{samplesize2014}. \citet{ribeiro-etal-2020-beyond} use the principles of software testing to test models, essentially behavioral testing the models using a \textit{CheckList} of test cases. The \textit{CheckList} tests the model against a set of  linguistic capabilities such as negation, replacing named entities etc. This requires careful curation of test examples and requires that these samples are updated as the capabilities of the model improve. Similar strategies have been employed to develop test suites for concept recognition systems \cite{cohen-etal-2010-test,groza-verspoor-2014-automated} and  negation inference \cite{truong-etal-2022-another}. \citet{kiela-etal-2021-dynabench} develop  Dynabench  to continuously update the test set samples with human-in-the-loop.

\subsection{Establishing cause and effect in models}
In clinical studies, randomized control trials (RCT) form the gold standard of evidence to establish \textit{cause and effect} of treatment or interventions \cite{Hariton2018-ti} and their outcomes. In a RCT, participants of the study are randomly assigned to a ``experimental" and ``control" group, where the experimental group receives the intervention and the control group receives a placebo  \cite{Kendall164}. The  key intuition is that if the only non-random difference between the experimental and control group is the intervention, then the intervention must be the cause of the outcome of the intervention. More specifically, causal effects can be measured by ``matching'' or balancing the distribution in the case and control groups, an approach that is used  in observational studies and RCTs to minimize bias or confounder effects \cite{Stuart2010-re,  PATERSON2024101889, 1975-06502-001}.

Adapting this approach to NLP benchmarks would involve curating a counterpart `control' test set for the standard randomly sampled  test set. The control test set would be created by making minor perturbations to a sample in the original test set, e.g.\  through the use of contrast sets \cite{gardner-etal-2020-evaluating}. The idea here is that if the model has effectively learned the key linguistic aspects required to predict a given label, then the model should also make the correct prediction when the key aspect is perturbed, see Table~\ref{tab:matched_pair_example}. In Table~\ref{tab:matched_pair_example}, some of the linguistic aspects that are measured are (a) meaning good vs bad (b) the impact of singular vs plural or swapping nouns on  sentiment.  Furthermore, we suggest that a model's prediction should be marked as correct if and only if its prediction on a perturbed counterpart is also correct.  This would ensure that the model is judged for its linguistic skills, evaluated in the matched pair, at least within the  context of the test set, ensuring internal validity. The  examples in Table~\ref{tab:matched_pair_example} are simplified to illustrate the key idea to measure causal effects, matching in NLP evaluation needs to be explored further.
\begin{table}[]
{\small
    \centering
    \begin{tabular}{c|c|c}
    \hline
     \textbf{Case (O)} & \textbf{Control (O)} & \textbf{S}\\ 
    \hline
         The movie is good. \ding{56} & The movie is bad. \ding{52}  & 0 \\
         The flower is pretty. \ding{52} & The flowers are pretty. \ding{56} & 0 \\
         Tom did a great job. \ding{52} & Jack did a  great job. \ding{52} & 1 \\
    \hline
    \end{tabular}
    }
    \caption{Simplified example of matched pairs  to measure causal effects. The outcome (O), or the model prediction,  can be either correct  \ding{52} or incorrect (\ding{56}), and the score (S) for a single test scenario is either 1 or 0. With matched pair evaluation, a model relying on spurious factors would be scored  $\frac{1}{3}=33.3\%$,  compared without matched pairs where each sample is treated independently $\frac{4}{6}=66.7\%$.}
    \label{tab:matched_pair_example}
\vspace{-6mm}
\end{table}

\subsection{How unreliable is a surface pattern?}\label{sec:unreliablesuracepattern}
While it may be impossible to prevent models from relying on surface patterns, we specifically need to watch  out for model's dependence  on spurious correlations or features that tend to be highly unreliable, e.g.\  the distance between the participating entities in relation extraction discussed above.  

Generalizability of deep learning networks depends on whether they learn \textbf{(a)}  spurious correlations, \textbf{(b}) reasonable heuristics, or \textbf{(c)} oracle true language meaning.  Oracle true meaning refers to true language understanding that takes into account the meanings of the individual words as well as the interplay between them relevant to 
a target task.
Spurious correlations are surface patterns with little or no linguistic backing tied to specific sample characteristics, whereas heuristics are plausible surface patterns that  may generally work without the need for deeper comprehension.
Generalizability is most adversely impacted by spurious correlations, making the model internally invalid. For instance, a model that associates the presence of the word ``good'' in a review with positive sentiment has captured a heuristic; it is a reasonable rule but may not produce a correct prediction when used in the context of 
negation or sarcasm. Spurious correlations can render the model  useless beyond the test set.  

\subsection{Replication studies and generalization}
Replication studies in psychology have brought the concerns of unreliable studies from scientific research to the forefront, including the possibility of spurious results to be accepted as genuine effects \cite{replicationstudies}. While  reproducibility of results in machine learning is a challenge \cite{Gundersen_Kjensmo_2018, belz-etal-2022-quantified}, replicating experiments in an independent dataset can help support or challenge the generalizability of an original finding \cite{Kukull2012-dr}.  For instance, if good model performance is only achieved in one test set and not replicable in any other tests, it points to possible issues in internal and/or external validity of the study.  As an example, \cite{mccoy-etal-2020-berts} identified that the performance of BERT  on the  original  MNLI \cite{mnli-williams-etal-2018-broad} test set is not consistently replicable on a  modified version of the test set, and investigations point to BERT relying on heuristics such as lexical overlap between the premise and hypothesis to achieve high scores on the official MNLI test set \cite{mccoy-etal-2019-right}. 

\subsection{Generalization in  large language models}
While pretrained and instruction fine-tuned ultra-large language models (LLMs) with billions of parameters  such as GPT-3 \cite{NEURIPS2020_1457c0d6} seem to demonstrate improved in-context zero or few shot generalization capabilities, compared to smaller models with a few hundred million parameters such as BERT  \cite{devlin-etal-2019-bert},  the jury is out on whether these results can be explained by improved memorization rather than generalization. This is in part because the official training data / test data from public datasets may not be fully independent – they could have been used to train such large models and the training data used is not publicly disclosed or well documented \cite{sainz-etal-2023-nlp, magar-schwartz-2022-data}. Furthermore, these LLMs tend to be highly sensitive to the prompts used, where characteristics that should \textit{not influence} a prompt’s interpretation  can result in  accuracy varying by over 25 points in LLMs including  GPT-4  and LLaMA-2-13B \cite{gan-mori-2023-sensitivity, sclar2024quantifying}. These facts bring into question the linguistic comprehension  of such LLMs.

Furthermore,  whether fine-tuning  such LLMs on  smaller target datasets with a few thousand training samples can achieve better generalization, compared to much smaller models like BERT, needs to be studied further. Regardless,  the need to verify that it is indeed linguistic capabilities that have resulted in the model’s performance and not the model relying on spurious correlations remains. This  requires ensuring that the test sets  are effective in measuring linguistic capabilities.

\section{Related works}

There are several reports of spurious correlations in SOTA  models and  benchmark datasets \citet{mccoy-etal-2019-right, gururangan-etal-2018-annotation, shinoda-etal-2022-look}, as discussed previously in this paper. \citet{gardner-etal-2021-competency} attempt to formalize the definition of which features can be deemed spurious. Approaches to detect spurious features without apriori handcrafting them  have also been studied, e.g.\ \citet{utama-etal-2020-towards}, relying on the observation that pre-trained models exploit simple patterns in the early stage of the training phase. Training a model to be robust against spurious correlations is also an active area of research,  e.g.\ \citet{tu-etal-2020-empirical} find that multitask learning  improves generalization. Data-driven approaches such as the use of h-adversarial training samples \cite{elangovan-etal-2023-effects}, appropriate  sample selection \cite{schwartz-stanovsky-2022-limitations}, or robust loss function \cite{he2023focused} can  improve model  robustness to spurious correlations. Causal deep learning and  inference can also assist in ensuring causality  \cite{Luo2020}.


\section{Conclusion}
Generalization  \underline{is an act of reasoning} that involves drawing broad inferences from specific observations \cite{Polit2010-lm}. While generalization of machine learning models in NLP is a complex topic, clinical research offers guiding principles  on how to understand generalization, including precise population definitions and contrasting controls. There are various points of failure in NLP modelling that can produce  results that may not be generalizable. Out-of-distribution (OOD) effects
may be an overly simplistic explanation 
for generalization failures.   Attributing  causation to generalization failures requires detailed investigation into some potential  pitfalls affecting internal validity. Ensuring internal validity requires at least an effective test set that 
controls for spurious correlations in 
the training data. More careful construction of evaluation frameworks will ensure appropriate inferences about the relationship between \textit{cause} (model's linguistic capabilities) and \textit{effect} (performance) from experimental results.

\section*{Limitations}

Internal validity can be compromised for various reasons, broadly speaking, data vs non-data related problems. Non data related issues can be bugs in code. Data related problems such label noise, poor quality annotation can also make a model internally invalid. In this paper, we have primarily focused on spurious correlations, assuming that the labels themselves are of high quality.

As mentioned  in section~\ref{sec:results}, identifying spurious correlations that a model relies on is non-trivial. A  simple surrogate model used in our case study can only detect the dominant spurious features, while a deep learning model such as BERT can learn thousands of such spurious features. Hence, if the simple surrogate model has poor performance, it does not mean that the model has not relied on  spurious features.

\section*{Ethics Statement}
This work does not involve collection of new data; all analysis relies on previously published data sets. Our work adheres to the ACL Ethics Policy\footnote{\url{https://www.aclweb.org/portal/content/acl-code-ethics}}.


\bibliography{references}



\clearpage
\newpage
\appendix
\onecolumn

\section{Appendix: PTM-PPI dataset Decision-tree}
\label{App:PTM-PPI dataset Decision-tree}
\begin{figure}[h]
    \centering
\includegraphics[width=1.0\linewidth]{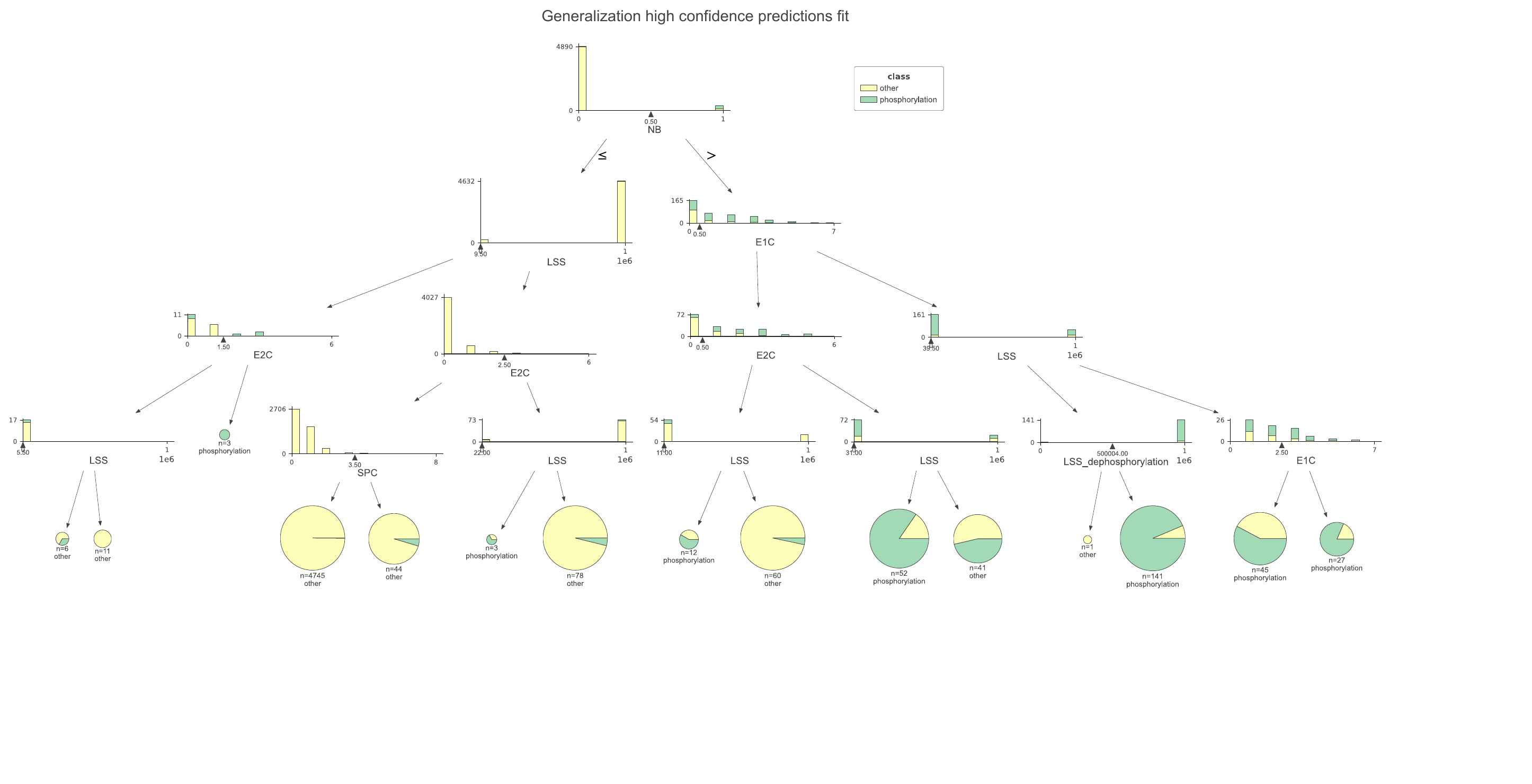}
    \caption{Decision Tree (NB-T) fit in high confidence predictions in the generalization set}
   \label{app:fig:decisiontreeGH}
\end{figure}

\begin{figure}[h]
    \centering
\includegraphics[width=1.0\linewidth]{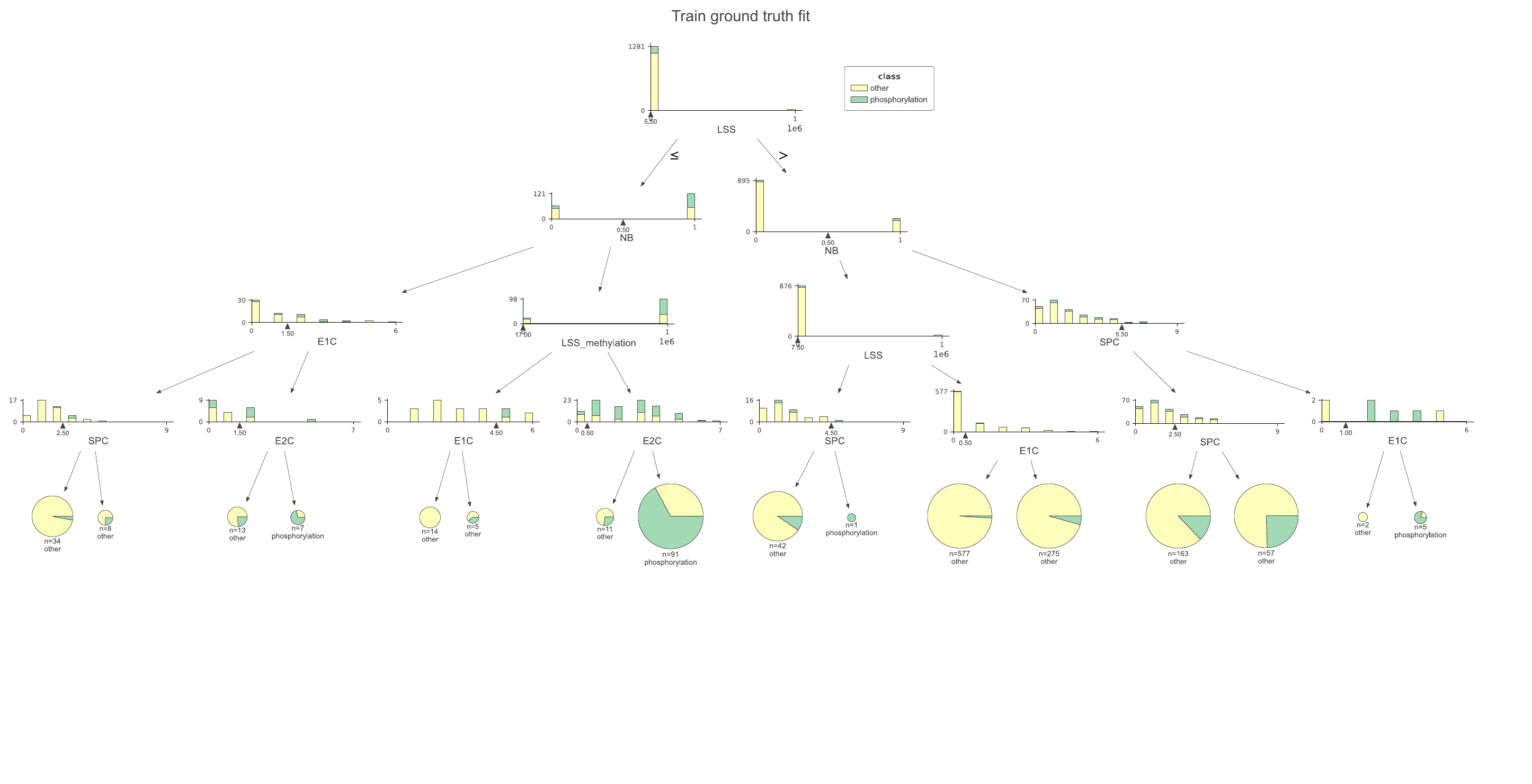}
    \caption{Decision Tree (NB-T) fit in Train ground truth fit}
   \label{app:fig:decisiontreeTrainGT}
\end{figure}

\begin{figure}[h]
    \centering
\includegraphics[width=1.0\linewidth]{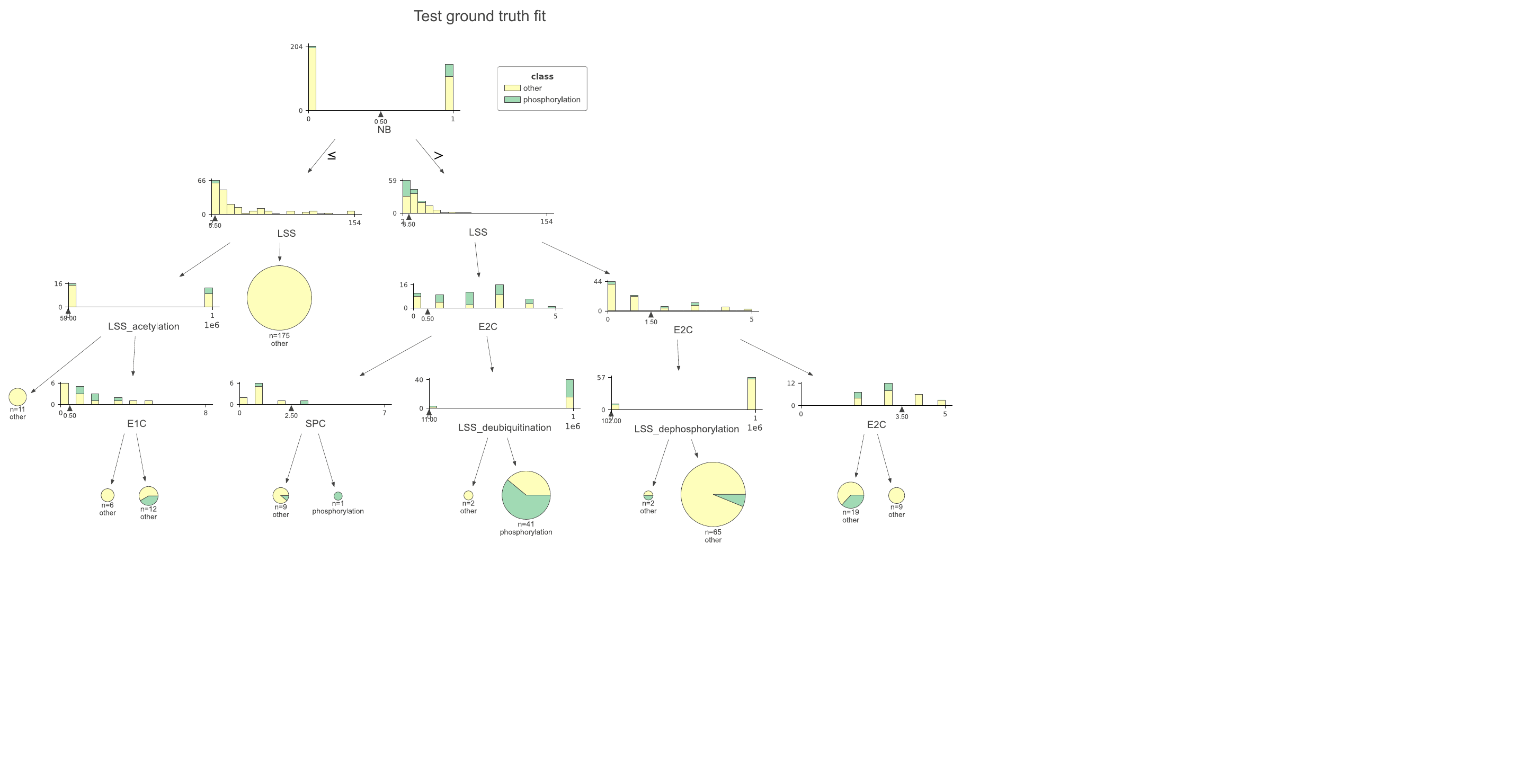}
    \caption{Decision Tree (NB-T) fit in Test ground truth fit}
   \label{app:fig:decisiontreeTestGT}
\end{figure}

\begin{figure}[h]
    \centering
\includegraphics[width=1.0\linewidth]{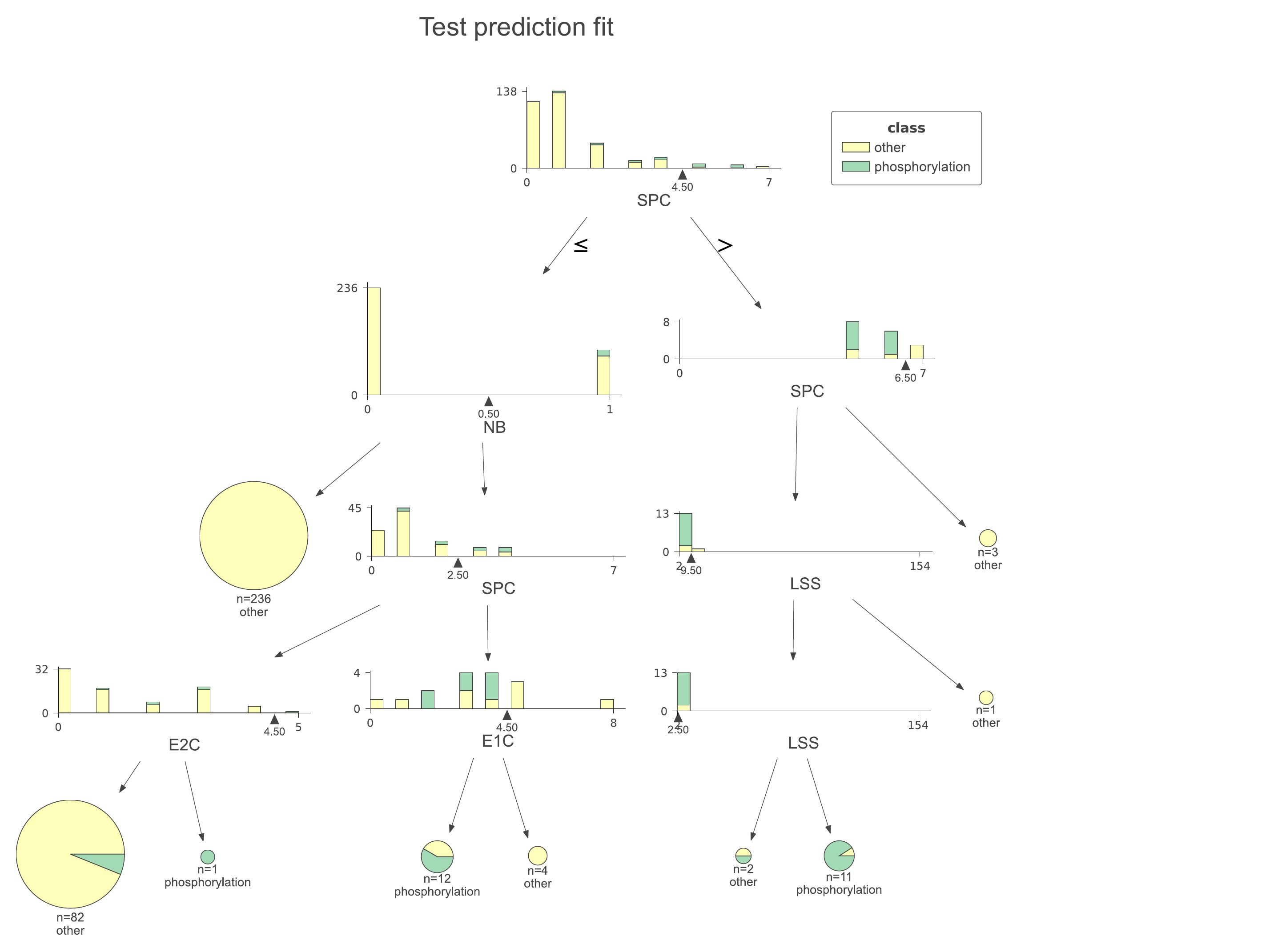}
    \caption{Decision Tree (NB-T) fit in Test set BioBERT predictions fit}
   \label{app:fig:decisiontreeTestPred}
\end{figure}

\clearpage
\newpage
\section{Appendix: SNLI Correlation}\label{app:sec:SNLI}
\begin{figure}[h]
    \centering
\includegraphics[width=1.0\linewidth]{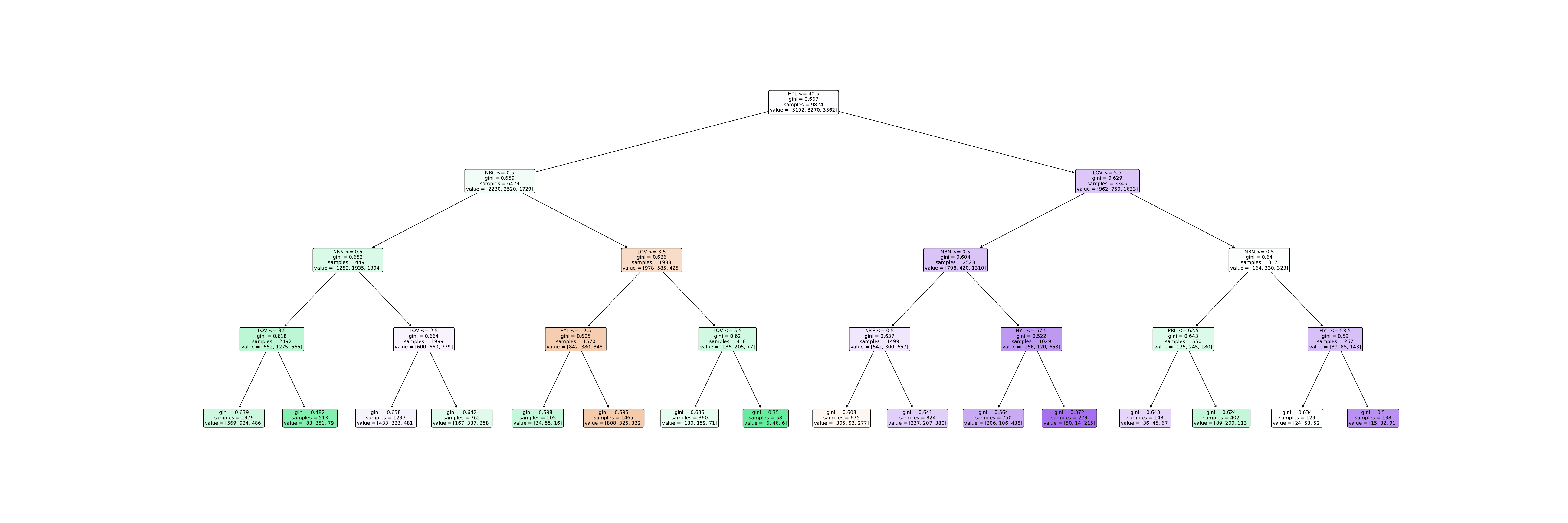}
    \caption{Decision Tree (NB-T) fit in SNLI Test set BERT predictions fit using HYP+PRM}
   \label{app:fig:decisiontreeBPSNLITestPred}
\end{figure}

\begin{figure}[h]
    \centering
\includegraphics[width=1.0\linewidth]{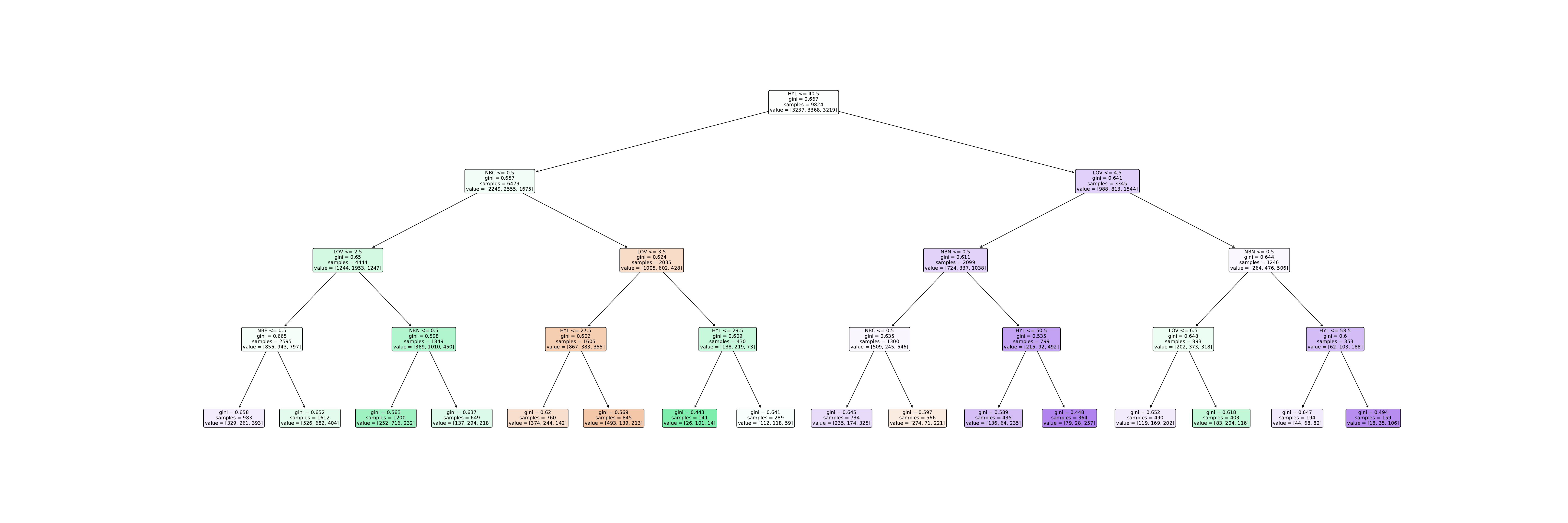}
    \caption{Decision Tree (NB-T) fit in SNLI Test set GT fit using HYP+PRM}
   \label{app:fig:decisiontreeGTSNLITestPred}
\end{figure}

We analyze the spurious correlations on the ground truth (GT) as well as the predictions (MP) from BERT \cite{devlin-etal-2019-bert} using features such as \textbf{(a)} the number of words (sentence length) of the hypothesis (HYL) or premise (PRL), \textbf{(b)} the presence of negation in the hypothesis (HNEG) and the premise (PNEG). As shown in Table~\ref{app:tab:SNLI}, using only the hypothesis achieves the highest $\kappa$ in the case of both train and test. We also find that hypothesis length (HYL) is a key feature appearing at the top of the   decision trees in Figure~\ref{app:fig:decisiontreeGTSNLITestPred} and Figure~\ref{app:fig:decisiontreeBPSNLITestPred}, which is also identified by \citet{gururangan-etal-2018-annotation}.

\begin{table}[h]
\begin{tabular}{lllr}
\hline
    Dataset  &  L & M & $\kappa$ \\
\hline
SNLI TR PRM & GT &    NB &  *0.00 \\
SNLI TR PRM & GT &   NB-T &  0.19 \\
SNLI TR HYP & GT &    NB &  0.25 \\
SNLI TR HYP & GT &   NB-T &  \textbf{0.27} \\
SNLI TR HYP+PRM & GT &    NB &  0.16 \\
SNLI TR HYP+PRM & GT &   NB-T &  0.22 \\
\hline
SNLI TS PRM & GT &    NB &  0.04 \\
SNLI TS PRM & GT &   NB-T &  0.20 \\
SNLI TS HYP & GT &    NB &  0.29 \\
SNLI TS HYP & GT &   NB-T &  \textbf{0.33} \\
SNLI TS HYP+PRM  & GT &    NB &  0.18 \\
SNLI TS HYP+PRM & GT &   NB-T &  0.24 \\
\hline
SNLI TS PRM & MP &    NB &  0.05 \\
SNLI TS PRM & MP &   NB-T &  0.20 \\
SNLI TS HYP & MP &    NB &  0.32 \\
SNLI TS HYP & MP &   NB-T &  \textbf{0.34} \\
SNLI TS HYP+PRM  & MP &    NB &  0.19 \\   
SNLI TS HYP+PRM  & MP &   NB-T &  0.25 \\
\hline
\end{tabular}  
\caption{Cohen's $\kappa$ on SNLI Train (SNLI TR) and Test (SNLI TS).   The Surrogate models NB and NB-T are used to predict the target label (L) -- the ground truth (GT) and the model prediction (MP) of BERT. The features used for Naive Bayes both during NB and stacked NB-T, can either be just the hypothesis (HYP) or just the premise (PRM) or both hypothesis and the premise (PRM).  * indicates that Cohen's kappa p-value is 0.10. For all other results, the p-value of Cohen's kappa is less than 0.05.}
\label{app:tab:SNLI}
\end{table}
 \end{document}